\documentclass[letterpaper]{article}
\usepackage{format_spec}
\usepackage[margin=1in]{geometry}
\usepackage{times}

\usepackage[hidelinks]{hyperref}
\usepackage[utf8]{inputenc}
\usepackage[small]{caption}
\usepackage{amsmath}
\usepackage{amsfonts}
\usepackage{amssymb}
\usepackage{graphicx}
\usepackage[colorinlistoftodos]{todonotes}
\usepackage{enumitem}
\usepackage{url}
\usepackage{array}
\usepackage{multicol}
\usepackage{tkz-graph}
\usepackage{mathtools}
\usepackage{bm}
\usepackage{algorithm}
\usepackage[noend]{algpseudocode}
\usepackage{color}
\usepackage{subcaption}
\usepackage[toc,page]{appendix}
\usepackage{booktabs}
\usepackage{dsfont}
\usepackage[round]{natbib}
\usepackage{varwidth}
\usepackage{mwe}

\newtheorem{theorem}{Theorem}[section]

\usepackage{xcolor}
\usepackage[framemethod=default]{mdframed} 
\mdfdefinestyle{default}{linecolor=gray, leftmargin=0, rightmargin=0, backgroundcolor=gray!40, innertopmargin=2pt, rightline=false, bottomline=false, leftline=false, topline=false}

\newcommand{\argmin}[1]{\underset{#1}{\operatorname{arg}\,\operatorname{min}}\;}

\newcommand{\thetab}{\boldsymbol{\theta}}
\newcommand{\thetabij}{\thetab_{ij}}
\newcommand{\nuisance}{\mathbf{u}}
\newcommand{\model}{\gen(\thetab,\nuisance)}
\newcommand{\X}{\mathbf{x}}

\newcommand{\xobs}{x^0}
\newcommand{\Xobs}{\X^0}

\newcommand{\Xsim}{\X_i}

\newcommand{\regionpair}{C_\epsilon}
\newcommand{\regioni}{C_\epsilon^i}
\newcommand{\regionuni}{U^i_\epsilon(\thetab)}
\newcommand{\optend}{\thetab^*_i}
\newcommand{\Doptend}{d^*_i}
\newcommand{\jac}{\mathbf{J}}
\newcommand{\hessian}{\mathbf{H}}

\newcommand{\optendname}{\text{optimisation end point}}   % In-text name for what I call "OMC sample"

\newcommand{\dete}{|\det (\A_i)|}
\newcommand{\determ}{\dete^{-1}}

\newcommand{\gpmodel}{\hat{d}_i}

\newcommand{\indicator}[1]{\mathds{1}_{#1}}
\newcommand{\sumstat}{\Phi}
\newcommand{\gensum}{\mathbf{f}}
\newcommand{\sumstatObs}{\mathbf{y}^0}
\newcommand{\E}{\mathop{\mathbb{E}}}

\newcommand{\A}{\mathbf{A}}
\newcommand{\gen}{g}
\newcommand{\voli}{\text{vol}(\regioni)}
\newcommand{\vole}{\text{vol}(B_{\epsilon})}
\newcommand{\w}{\mathbf{w}}
\newcommand{\hpost}{\bar{h}} % \E[h(\thetab) | \Xobs]
\newcommand{\ESS}{\textrm{ESS}}

\newcommand{\tran}{\mkern-2mu\raise1.25ex\hbox{$\scriptscriptstyle\top$}\mkern-3.5mu}
\newcommand{\tranlow}{\mkern-2mu\raise0.75ex\hbox{$\scriptscriptstyle\top$}\mkern-3.5mu}

\newcommand{\algformat}[1]{\textbf{#1}}

\title{Robust Optimisation Monte Carlo}

\author{Borislav Ikonomov\vspace{3mm}\\ 
	School of Informatics\\
	University of Edinburgh\\
	\textit{borislav.ikonomov@ed.ac.uk}\And
	Michael U. Gutmann\vspace{3mm}\\ 
	School of Informatics\\
	University of Edinburgh\\
	\textit{michael.gutmann@ed.ac.uk}
}

\begin{document}

\maketitle

%%%%%%%%%%%%%%%%%%%%%%%%%%%%%%%%%%%%%%%%%%%%%%%%%%%%%%%%%%%%
%%%%%%%%%%%%%%%%%%%%%%% Abstract %%%%%%%%%%%%%%%%%%%%%%%
%%%%%%%%%%%%%%%%%%%%%%%%%%%%%%%%%%%%%%%%%%%%%%%%%%%%%%%%%%%%
\begin{abstract}
    This paper is on Bayesian inference for parametric statistical models that are  defined by a stochastic simulator which specifies how data is generated. Exact sampling is then possible but evaluating the likelihood function is typically prohibitively expensive. Approximate Bayesian Computation (ABC) is a framework to perform approximate inference in such situations. While basic ABC algorithms are widely applicable, they are notoriously slow and much research has focused on increasing their efficiency. Optimisation Monte Carlo (OMC) has recently been proposed as an efficient and embarrassingly parallel method that leverages optimisation to accelerate the inference. In this paper, we demonstrate an important previously unrecognised failure mode of OMC: It generates strongly overconfident approximations by collapsing regions of similar or near-constant likelihood into a single point. We propose an efficient, robust generalisation of OMC that corrects this. It makes fewer assumptions, retains the main benefits of OMC, and can be performed either as post-processing to OMC or as a stand-alone computation. We demonstrate the effectiveness of the proposed Robust OMC on toy examples and tasks in inverse-graphics where we perform Bayesian inference with a complex image renderer.
\end{abstract}

%%%%%%%%%%%%%%%%%%%%%%%%%%%%%%%%%%%%%%%%%%%%%%%%%%%%%%%%%%%%
%%%%%%%%%%%%%%%%%%%%%%% Introduction %%%%%%%%%%%%%%%%%%%%%%%
%%%%%%%%%%%%%%%%%%%%%%%%%%%%%%%%%%%%%%%%%%%%%%%%%%%%%%%%%%%%
\section{INTRODUCTION}

Simulator-based models can describe many complex processes that occur in nature, such as the evolution of genomes \citep{sim_models_genome_evolution} or the dynamics of gene regulation \citep{sim_models_gene_dynamics}. Learning their parameters, in particular when done in a Bayesian framework, allows us to make predictions or take decisions based on incomplete information. However, learning the parameters or obtaining their posterior distribution is typically computationally very demanding as their likelihood functions are intractable. Likelihood-Free Inference (LFI) methods have thus emerged that perform inference when the likelihood function is not available in closed form but sampling from the model is possible.

A prominent instance of LFI is Approximate Bayesian Computation (ABC); for recent reviews, see for example \citep{Sisson2018,abc_fundamentals}. Other instances of LFI are the synthetic likelihood approach by \citet{Wood2010} and its generalisations \citep{Thomas2016, Price2017, Fasiolo2018}. This paper focuses on ABC where the basic idea is to identify the parameter values which generate synthetic data that is close to the observed data under some chosen discrepancy measure. This measure can be the Euclidean distance between suitably chosen summary statistics, but other measures are possible too \citep[e.g.][]{abc_classification_as_distance, Bernton2018}. Generally, there are two main avenues of research for ABC --- one focuses on improving the distance metric and/or the summary statistics used \citep[e.g.\ ][]{abc_summary_stats_aeschbacher, Fearnhead2012}, while the other concentrates on computational efficiency \citep[e.g.\ ][]{abc_regression_abc_beaumont, Blum2013, omc, abc_bolfi, abc_epsilonfree_with_bayesian_cde}. This paper focuses on the latter, and as such we assume the distance and summary statistics are given.

The primary focus of this paper is Optimisation Monte Carlo (OMC) --- an ABC method developed by \citet{omc} and also independently by \citet{reverse_sampler, reverse_sampler_2} under the name of ``the reverse sampler''. It uses optimisation to efficiently produce weighted posterior samples in a fully parallelisable manner, which makes it a desirable ABC method.

\begin{figure}[ht]
	\centering
	\includegraphics[width=0.95\linewidth]{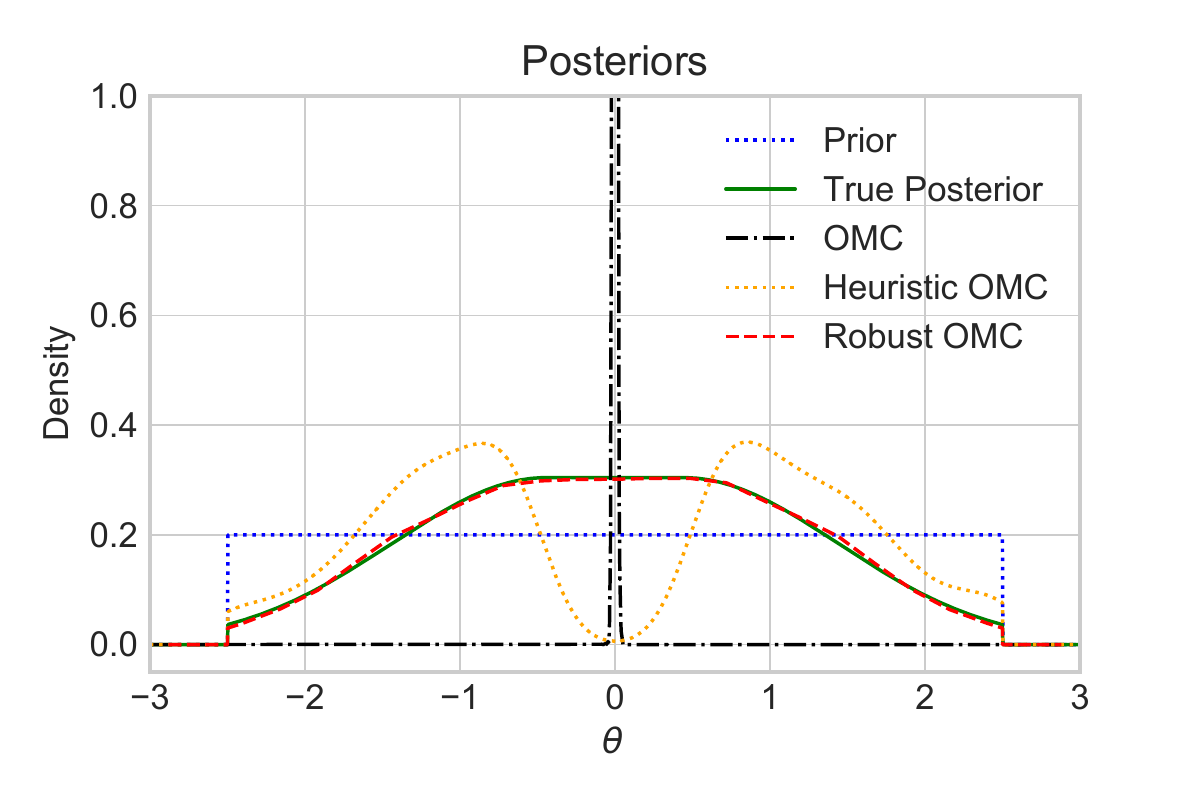}
	\caption{\label{fig:omc_fail}An example where OMC fails to approximate the true posterior, collapsing a region of similar likelihood into a single point. Heuristic OMC is a simple heuristic that (unsuccessfully) attempts to solve this issue. Robust OMC is the approach proposed in this paper. See Subsection~\ref{subsec:exp_1} for details.}
\end{figure}

A weight produced by OMC represents the volume of the parameter region around a posterior sample which contains points that are as good as the sample. These points should thus be considered to be samples from the posterior too. However, if this region is particularly big, it is no longer appropriate to approximate the entire region with a single point and, as a result, OMC produces an overly confident posterior. Figure~\ref{fig:omc_fail} illustrates this failure case for a simple 1D scenario. We can see that OMC fails to characterise the posterior uncertainty and collapses regions of similar likelihood into a single point.

We propose Robust OMC (ROMC) which efficiently identifies the regions of acceptable points themselves and samples from them directly. These regions are characterised by the set of all parameters which generate data whose distance to the observed data is less than a certain threshold. Instead of approximating a possibly very large region with a single point, we draw samples from the region which then replace the original OMC sample along with its weight.
The main improvements of ROMC over OMC are:\vspace{-1ex}
\begin{itemize}
	\item It handles likelihoods that are (nearly) flat on significant regions in parameter space. All our experiments confirm this.\vspace{-0.5ex}
	\item It can be applied as post-processing to an original OMC run (see experiment 2), or as part of a stand-alone run (see experiment 3).\vspace{-0.5ex}
		%either as part of the initial OMC computations, or entirely as post-processing. The latter is the case in our experiment 2, and the former in experiment 3.
	\item OMC requires that the derivatives of the simulator can be computed or reasonably approximated, while ROMC does not (see experiment 3).% To showcase this, experiment 3 is done without the use of simulator derivatives.
	% \item OMC further requires the use of the Euclidean distance between the summary statistics of the simulated and observed data. In the proposed approach, any distance function will do.
\end{itemize}
\vspace{-2ex}
%The rest of the paper is structured as follows. Section~\ref{sec:background} covers the basics of both Rejection ABC and OMC, and goes into more detail about the OMC failure case illustrated in Figure~\ref{fig:omc_fail}. In Section~\ref{sec:romc}, we derive the main equations that govern Robust OMC, show under what assumptions it becomes equivalent to OMC, and present our practical implementation of it. Section~\ref{sec:exp} empirically demonstrates several cases where OMC fails to produce a reasonable approximation to the posterior, but Robust OMC succeeds. Section~\ref{sec:conc} concludes the paper.

%%%%%%%%%%%%%%%%%%%%%%%%%%%%%%%%%%%%%%%%%%%%%%%%%%%%%%%%%%%%
%%%%%%%%%%%%%%%%%%%%%%% ABC Sampling %%%%%%%%%%%%%%%%%%%%%%%
%%%%%%%%%%%%%%%%%%%%%%%%%%%%%%%%%%%%%%%%%%%%%%%%%%%%%%%%%%%%
\section{BACKGROUND}\vspace{-1.5ex}
\label{sec:background}
We here present the basics of Approximate Bayesian Computation (ABC), review the OMC algorithm by \citet{omc}, and discuss when it collapses regions of similar likelihood into a single point.
\subsection{Rejection ABC}

ABC methods produce samples from an approximate posterior \citep[see e.g.][]{abc_fundamentals}. They generally make minimal assumptions about the model and only assume black-box access to the simulator $\model$. The target parameters $\thetab$, which we wish to infer, are used as input to the simulator which then stochastically produces synthetic data by drawing $\nuisance$ using a random number generator. These $\nuisance$ correspond to nuisance variables since we do not aim to find a distribution over them. %ABC further requires a distance function that captures the difference $d(\X, \Xobs)$ between the simulated data $\X$ and the observed data $\Xobs$. 

The simplest ABC method is Rejection ABC. In each iteration, data $\Xsim$ is simulated using the generative model $\gen(\thetab, \nuisance_i)$ for some setting $\nuisance_i$ of the nuisance variables and with parameter values sampled from the prior. The distance between the simulated and observed data $d_i = d(\Xsim, \Xobs)$ is then computed and stored. After a sufficiently large amount of samples has been generated, the algorithm accepts those having the $n$ lowest $d_i$ as samples from the approximate posterior. See Algorithm~\ref{alg:rej_abc} in the appendix for pseudo-code. 

While simple and robust, Rejection ABC is known to be computationally inefficient, especially when the prior space is large \citep[e.g.\ ][]{abc_fundamentals}. In most cases, more sophisticated methods are necessary.

\subsection{Optimisation Monte Carlo}
\label{subsec:background_omc}
We start our brief review of Optimisation Monte Carlo (OMC) by noting that ABC algorithms in general implicitly approximate the likelihood function by the probability $\Pr(d(\gen(\thetab, \nuisance), \Xobs) \le \epsilon)$ that the generated data is within distance $\epsilon$ of the observed data \citep[e.g.][]{abc_fundamentals}. ABC algorithms thus produce samples from the following approximate posterior 
\begin{align}
p^*_\epsilon(\thetab | \Xobs) &\propto p(\thetab) \Pr(d(\gen(\thetab, \nuisance), \Xobs) \le \epsilon | \thetab)\\
&\propto p(\thetab) \int p(\nuisance) \mathds{1}_{\regionpair}(\thetab, \nuisance)\ d\nuisance,
\label{eq:abc_post}
\end{align}
where $p(\nuisance)$ is the density of the nuisance variables $\nuisance$, $\regionpair = \{(\thetab, \nuisance) : d(\gen(\thetab, \nuisance), \Xobs) \le \epsilon\}$ is the set of points
$(\thetab, \nuisance)$ for which the distance is below the threshold, and $\indicator{\regionpair}(\thetab, \nuisance)$ is an indicator function that equals one only if $(\thetab, \nuisance) \in \regionpair$. While this formulation uses the indicator function (boxcar kernel), more general kernels can be used as well.

The integral over $\nuisance$ corresponds to an expectation with respect to $p(\nuisance)$ and can thus be approximated as a sample average so that we obtain the approximation
\begin{align}
p_\epsilon(\thetab | \Xobs) &\propto  p(\thetab) \frac{1}{n} \sum_{i=1}^n \mathds{1}_{\regionpair}(\thetab, \nuisance_i),  \label{eq:abc_base}
\end{align}
where the $\nuisance_i$ are sampled from $p(\nuisance)$. Importantly, this formulation essentially removes the randomness from the simulator: $\gen(\thetab, \nuisance_i)$, which occurs in each $\indicator{\regionpair}(\thetab, \nuisance_i)$ by definition of $\regionpair$, is a deterministic function of $\thetab$ because $\nuisance_i$ is held fixed.

OMC exploits the fact that $\gen(\thetab, \nuisance_i)$ is a deterministic function in order to accelerate the sampling from the posterior in \eqref{eq:abc_base}. For each $\nuisance_i$, OMC finds a value $\optend$ for which $\gen(\optend, \nuisance_i)$ and $\Xobs$ are within distance $\epsilon$. Importantly, this is done by minimising the deterministic cost function $d(\gen(\thetab, \nuisance_i), \Xobs)$ with respect to $\thetab$. Note that settings of $\nuisance_i$ for which no $\optend$ has distance below $\epsilon$ are excluded from the sum and thus do not affect the posterior.

\citet{omc} consider the case where the distance $d(\gen(\thetab, \nuisance),  \Xobs))$ is the Euclidean distance between some summary statistics $\sumstat$ of the generated and observed data. We denote the summary statistics of the observed data by $\sumstatObs = \sumstat(\Xobs)$ and we further absorb the computation of the summary statistics into the simulator so that we obtain $\gensum(\thetab, \nuisance) = \sumstat(\gen(\thetab, \nuisance))$, which can be regarded as a generative model on the level of the summary statistics. With this notation, OMC considers the distance $d(\gen(\thetab, \nuisance), \Xobs) = ||\gensum(\thetab, \nuisance) - \sumstatObs||$.

OMC approximates the posterior $p^*_\epsilon$ in the limit of $\epsilon \to 0$ as a mixture of weighted point masses centred at the minimisers $\optend$:  $p(\thetab | \Xobs) \propto \sum_{i=1}^{n} w_i \delta(\thetab-\optend)$. The value $w_i$ is a weight that reflects the local behaviour of the distance function and hence $\gensum(\thetab, \nuisance_i)$ around $\optend$. As shown by \citet{omc}, it equals $ p(\optend) * \det(\jac_i\tran \jac_i)^{-1/2}$ where $\jac_i$ is the Jacobian matrix with columns $\partial \gensum(\optend, \nuisance_i) / \partial \theta_k$ with $\theta_k$ denoting the $k$-th element of $\thetab$.  Algorithm~\ref{alg:omc} summarises the OMC algorithm. For further details, we refer the reader to the original paper by \citet{omc}.

\begin{algorithm}
	\caption{Optimisation Monte Carlo. Generates $n$ independent samples $\optend$ with weights $w_i$ from the approximate posterior. %Needs summary statistics simulator %$\gensum(\thetab, \nuisance)$ and observed summary statistics $\sumstatObs$.
	% commented out the details because the other algorithms don't have it either + it saves space
	}\label{alg:omc}
	\begin{algorithmic}[1]
		\For{$i \gets 1 \textrm{ to } n$}
		\State $\nuisance_i \sim p(\nuisance)$ \Comment{Draw nuisance parameters $\nuisance_i$.}
		\State $\optend = \argmin{\thetab}  ||\gensum(\thetab, \nuisance_i) - \sumstatObs||$ \Comment{Optimisation.}
		\State Compute $\jac_i$ with columns $\partial \gensum(\optend, \nuisance_i) / \partial \theta_k$
		\State Compute $w_i = p(\optend) * (\det(\jac_i \tran \jac_i))^{-1/2}$
		\State Accept $\optend$ as posterior sample with weight $w_i$.
		\EndFor
	\end{algorithmic}
\end{algorithm}

As can readily be seen from Algorithm~\ref{alg:omc}, ill-conditioned matrices $\jac_i \tran \jac_i$ produce very large weights for the corresponding $\optend$, possibly completely overshadowing the remaining samples and creating an approximate posterior density that is spiked at a single location (as in Figure~\ref{fig:omc_fail}). One may think that this issue can be easily fixed by regularising $\jac_i \tran \jac_i$ before computing the determinant. However, the issue goes deeper: ill-conditioned matrices $\jac_i \tran \jac_i$ occur when a large parameter region around the optimum $\optend$ produces data with small distances. These regions are poorly approximated by point masses or infinitesimally small ellipsoids, and amending the value of the weight cannot correct for this.\footnotemark In other words, the OMC failure mode occurs when a large parameter region around $\optend$ is a solution to $||\gensum(\thetab, \nuisance_i) - \sumstatObs|| \le \epsilon$, which happens, for example, when the likelihood function is (nearly) constant around $\optend$.

\footnotetext{As pointed out in the original paper as a limitation of OMC, this happens if there are fewer summary statistics than parameters. But as shown here, this failure mode is more general. The proposed robust method provides a solution.}

%%%%%%%%%%%%%%%%%%%%%%%%%%%%%%%%%%%%%%%%%%%%%%%%%%%%%%%%%%%%
%%%%%%%%%%%%%%%%%%%%%%% Robust OMC %%%%%%%%%%%%%%%%%%%%%%%
%%%%%%%%%%%%%%%%%%%%%%%%%%%%%%%%%%%%%%%%%%%%%%%%%%%%%%%%%%%%
\section{ROBUST OPTIMISATION MONTE CARLO}
\label{sec:romc}

We here develop a framework and concrete algorithms that have the benefits of OMC but do not collapse areas of similar likelihood into a point-mass. %The following subsections present the general theory, the conditions under which it becomes equivalent to OMC, and a specific implementation. % omitted to save space

\subsection{The Robust OMC Framework}
We start from the basic characterisation of the finite sample version of the ABC posterior in Equation \eqref{eq:abc_base} which holds irrespective of OMC. Under this approximation, the expectation of an arbitrary function $h(\thetab)$ under the posterior $p_\epsilon(\thetab | \Xobs)$ is
\begin{align}
\E[h(\thetab) | \Xobs] =& \frac{ \int h(\thetab) p(\thetab) \frac{1}{n} \sum_{i=1}^n \indicator{\regionpair}(\thetab, \nuisance_i)\ d \thetab}{\int p(\thetab) \frac{1}{n} \sum_{i=1}^n \indicator{\regionpair}(\thetab, \nuisance_i)\ d \thetab}\\
=& \frac{\sum_{i=1}^{n} \int h(\thetab) p(\thetab)  \indicator{\regioni}(\thetab) \ d\thetab}{\sum_{i=1}^{n} \int p(\thetab) \indicator{\regioni}(\thetab) \ d\thetab},
\label{eq:romc_base}
\end{align}
where $\regioni = \{\thetab : d(\gen(\thetab, \nuisance_i), \Xobs) \le \epsilon\}$ is the set of parameters where, for a particular random seed or realisation of $\nuisance_i$, the simulated data is within distance $\epsilon$ from the observed data. %This means that in the integrals above, the set $\regionpair$ containing tuples ($\thetab,\nuisance)$ is %traversed across the $\thetab$-dimension for a fixed value of $\nuisance$ while in the integral in \eqref{eq:abc_post}, %it is traversed along the $\nuisance$-dimension for a fixed value of $\thetab$. 
Equation \eqref{eq:romc_base} features $n$ integrals $I_i$ in the numerator,
\begin{equation}
    I_i =  \int h(\thetab) p(\thetab) \indicator{\regioni}(\thetab) \ d\thetab,
\end{equation}
and similar ones in the denominator.  The integrals are generally intractable but since they correspond to an expectation with respect to the prior $p(\thetab)$, they could be approximated by a sample-based average because sampling from the prior is typically possible in likelihood-free inference problems. However, this would be inefficient in the case of a broad prior as most samples would give $\indicator{\regioni}(\thetab)=0$, i.e.\ they would essentially get rejected much like in rejection ABC. It is more efficient to sample from a proposal distribution $q_i(\thetab)$ that only has support on the acceptance region $\regioni$. We will discuss how to construct such $q_i(\thetab)$ in Subsection~\ref{subsec:romc_impl}. Assuming we have a suitable $q_i(\thetab)$, the integrals $I_i$ can be approximated as
\begin{align}
I_i    & = \int h(\thetab) \indicator{\regioni}(\thetab) \frac{p(\thetab)}{q_i(\thetab)} q_i(\thetab) \ d\thetab\\
   %  & = \E_{\thetab \sim q_i} \left[ h(\thetab) \indicator{\regioni}(\thetab) \frac{p(\thetab)}{q_i(\thetab)}\right]\\
     & \approx \frac{1}{m} \sum_{j=1}^{m} h(\thetabij) \indicator{\regioni}(\thetabij) \frac{p(\thetabij)}{q_i(\thetabij)} ,
\end{align}
where $\thetabij \sim q_i(\thetab)$, and equivalently for the integral in the denominator. Replacing the integrals in \eqref{eq:romc_base} with their sample-based approximations, we obtain
\begin{align}
\E[h(\thetab) | \Xobs] \approx&  \frac{ \sum_{i=1}^{n} \sum_{j=1}^{m} h(\thetabij) \indicator{\regioni}(\thetabij) \frac{p(\thetabij)}{q_i(\thetabij)}}{\sum_{i=1}^{n} \sum_{j=1}^{m} \indicator{\regioni}(\thetabij) \frac{p(\thetabij)}{q_i(\thetabij)}}.\label{eq:romc_eff}
\end{align}
This expression corresponds to a weighted sample average of $h(\thetab)$. Denoting $\E[h(\thetab) | \Xobs]$ by $\hpost$, we have
\begin{align}
\hpost \approx&  \frac{\sum_{ij} w_{ij}  h(\thetabij)}{\sum_{ij} w_{ij}},&
w_{ij}  =& \indicator{\regioni}(\thetabij) \frac{p(\thetabij)}{q_i(\thetabij)}, \label{eq:weight}
\end{align}
where the $w_{ij}$ are the (unnormalised) weights and the samples $\thetabij$ are drawn from $q_i(\thetab)$. Since our test function $h(\thetab)$ has been arbitrary, this means that to obtain samples from the approximate posterior, we first draw $n$ samples $\nuisance_i$,\footnote{In practice, this is done by fixing the seeds of the simulator.} thus defining the acceptance region $\regioni$, and then $m$ samples $\thetabij$ from the corresponding proposal distribution $q_i(\thetab)$. This process is what we refer to as the Robust OMC approach. 

Before discussing the construction of the proposal distributions $q_i(\thetab)$, we show how OMC is obtained from \eqref{eq:romc_eff} by making additional assumptions. Some of the assumptions can be easily violated in practice, which then leads to the failure mode pointed out above and illustrated in Figure \ref{fig:omc_fail}.

\subsection{Connection to OMC}
We present here the assumptions under which the proposed Robust OMC (ROMC) approach becomes standard OMC. It shows that ROMC is both more general and more robust than standard OMC. 
\begin{theorem}
\label{theo:omc}

Under the below assumptions, ROMC becomes equivalent to standard OMC as $\epsilon \to 0$.

\textbf{Assumption 1.} The distance $d(\gen(\thetab, \nuisance), \Xobs)$ is given by the Euclidean distance between summary statistics $||\gensum(\thetab, \nuisance) - \sumstatObs||$.

\textbf{Assumption 2.} The proposal distribution $q_i(\thetab)$ is the uniform distribution on $\regioni$.

\textbf{Assumption 3.} The acceptance regions $\regioni$ are approximated by the ellipsoid $\regioni = \{\thetab : (\thetab - \optend)\tran \jac_i\tran\jac_i (\thetab - \optend) \leq \epsilon\}$ where $\jac_i$ is the Jacobian matrix with columns $\partial \gensum(\optend, \nuisance_i) / \partial \theta_k$.

\textbf{Assumption 4.} The matrix square root $\A_i$ of $\jac_i\tran\jac_i$ is full rank, i.e.\ $\text{rank}(\A_i)=\text{dim}(\thetab)$. 

\textbf{Assumption 5.} The prior is constant on the acceptance regions $\regioni$, i.e.\ $p(\thetab) = p(\optend)$ for $\thetab \in \regioni$. 
\end{theorem}

The proof of Theorem~\ref{theo:omc} is given in the appendix.

Assumptions 1 and 2 are of technical nature, but Assumption 1 highlights that ROMC can also use distances other than Euclidean ones. Assumption 3 and 4 show that OMC relies on $\regioni$ being well approximated by an ellipsoid of finite volume whose shape is determined by the local behaviour of $\gensum(\thetab, \nuisance_i)$ at $\optend$. The failure case described in Subsection~\ref{subsec:background_omc} and illustrated in Figure \ref{fig:omc_fail} is caused by a violation of these two assumptions. Assumption 5 is also important because it shows that e.g.\ strong smoothing of the empirical distribution defined by the weighted samples in OMC would ignore that the prior distribution may not be constant on the corresponding finite-sized ellipsoid. 

\subsection{Robust OMC Algorithms}
\label{subsec:romc_impl}
The Robust OMC (ROMC) framework has three key ingredients: the optimisation procedure as in OMC, the specification of the $\epsilon$-threshold as usual in ABC, and the proposal distribution. We here consider two sets of choices for these ingredients, resulting in Algorithms~\ref{alg:romc_box} and~\ref{alg:romc_ell} detailed below. The former algorithm assumes access to (approximate) simulator gradients and can be run as post-processing to standard OMC, and the latter is gradient-free. The two algorithms show that the proposed ROMC framework is versatile and that it can be used to exploit specific properties of the model.

%There are three specific elements that are of particular interest - the optimisation procedure, the construction of the proposal distribution, and the specification of the $\epsilon$-threshold. Note that all of these can be freely customised by the user to suit their specific needs --- our implementation below is not the only possibility, although we believe it is robust enough to be applicable in a wide variety of situations.

\subsubsection{Optimisation Step}

To obtain the $\optendname$ $\optend$, any optimisation algorithm can be used as long as it can minimise the distance with respect to $\thetab$.

\algformat{Algorithm~\ref{alg:romc_box}:} If gradients of the simulator are available, standard gradient-based optimisers are applicable.

\algformat{Algorithm~\ref{alg:romc_ell}:} If the simulator gradients are not available, we propose using Bayesian optimisation, which is a powerful optimisation scheme for objective functions that can be evaluated but whose gradients are not available \citep[see e.g.\ ][]{Shahriari2016}. In the simulations below, we use standard Bayesian optimisation (GPyOpt with default settings) that builds a Gaussian Process surrogate model $\gpmodel$ for each distance $d(\gen(\thetab, \nuisance_i), \Xobs)$ that needs to be minimised. The main purpose of the surrogate model in Bayesian optimisation is to decide at which $\thetab$ to evaluate the distance next. This also applies to our situation but for ROMC, there are two further uses of the surrogate model: 1) it can be used to greatly speed up the acceptance check $\indicator{\regioni}(\thetabij)$ in Equation~\eqref{eq:weight} by using the surrogate distance rather than the true distance (see experiment 3); and 2) it can facilitate the construction of the proposal distribution $q_i(\thetab)$ as discussed below.

\subsubsection{Threshold}

\algformat{Algorithms~\ref{alg:romc_box} and \ref{alg:romc_ell}:} ROMC requires a value for the threshold $\epsilon$ that occurs in the term $\indicator{\regioni}(\thetabij)$. This requirement to set a threshold is similar to most ABC algorithms where it is typically chosen as a small quantile of the observed distances (for other solutions see e.g.\ the work by \citet{abc_regression_abc_beaumont, Blum2010, abc_epsilonfree_with_bayesian_cde, Chen2019, Simola2019}). We take a similar approach but base the value of $\epsilon$ on the distances $\Doptend$ at the $\optendname$s, $\Doptend = d(\gen(\optend, \nuisance_i), \Xobs)$. Since the $\Doptend$ are the minimal distances obtained in the optimisation step, their values are much smaller than the distances that one would see in other ABC algorithms, and we can choose a large quantile. In our simulations we chose the 90\% quantile of the $\Doptend$ in order to be robust against bad optimisation instances. Since all $\thetabij$ are saved, the exact value for $\epsilon$ can be changed later by the user without incurring any overheads. 

\begin{algorithm}[t]
	\caption{Boxed Robust OMC. Requires simulator gradients, possible as post-processing to standard OMC.}\label{alg:romc_box}
	\begin{algorithmic}[1]
		\For{$i \gets 1 \textrm{ to } n$}
		\State Obtain $\optendname$ $\optend$.
	%	\State Choose search directions. Move along each one until a point $\thetab^{end}$ is reached for which $d(\gen(\thetab^{end}, \nuisance_i), \Xobs) < \epsilon_{big}$ no longer holds.
%		\State Create a box which has all $\thetab^{end}$ on its edges.
        \State Use curvature of $\jac_i\tran \jac_i$ to create a bounding box with volume $V_i$ as described in Subsection~\ref{subsec:prop_dist}.
		\State Define a uniform distribution $q_i(\thetab)$ over the box.
			\For{$j \gets 1 \textrm{ to } m$}
			\State $\thetabij \sim q_i(\thetab)$
			\State \begin{varwidth}[t]{0.9\linewidth}Accept $\thetabij$ as posterior sample with weight $w_{ij} = \indicator{\regioni}(\thetabij) * p(\thetabij) * V_i$\end{varwidth}
			\EndFor
		\EndFor
	\end{algorithmic}
\end{algorithm}

\subsubsection{Proposal Distribution}
\label{subsec:prop_dist}

We here describe two methods to construct the proposal distributions $q_i(\thetab)$. We assume that the optimisation step has given us a sample $\optend$ that is within $\regioni$. 

\algformat{Algorithm~\ref{alg:romc_box}:} If simulator gradients are available, then it is possible to compute the matrix $\jac_i\tran \jac_i$. Its eigenvectors are orthogonal directions of highest curvature, along which we scan until we reach a point whose resulting distance no longer falls under the threshold. Doing so in each dimension specifies a box, and defining a uniform distribution on this box gives us the proposal distribution $q_i(\thetab)$. Since $\jac_i\tran \jac_i$ is computed by standard OMC, this approach can be done entirely as post-processing to it.

\begin{figure}[h!]
	\centering
	\begin{subfigure}{.445\textwidth}
		\centering
		\includegraphics[width=1.\linewidth]{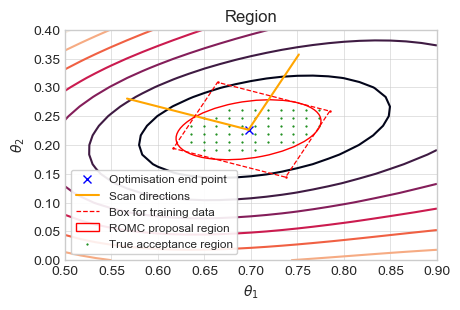}
	\end{subfigure}
	\caption{\label{fig:implementation_example}Algorithm~\ref{alg:romc_ell}, example proposal distribution $q_i(\thetab)$. The contours show the GP model distance, the green dots visualise the true acceptance region $\regioni$, and $q_i(\thetab)$ is the uniform distribution on the red ellipse which well approximates $\regioni$.}
\end{figure}

\algformat{Algorithm~\ref{alg:romc_ell}:}
Without simulator gradients, we construct a box around the optimisation end point as in Algorithm 2, except that we use the Hessian of the GP model instead of $\jac_i\tran \jac_i$. To robustify the approach against e.g.\ inaccuracies in the GP model and hence estimation of the curvature, we sample parameter values from inside the box and compute their distance to the observed data using the posterior mean of the GP model. This incurs practically no overhead and is considerably faster than using the true distances if the simulator is expensive. These parameter-distance pairs are then used to train a quadratic regression model of the distance. The contour where the distance is equal to the threshold defines an ellipsoid, and we use the uniform distribution on it as proposal distribution $q_i(\thetab)$. Figure \ref{fig:implementation_example} visualises the construction.

The appendix has further details on both algorithms.

\begin{algorithm}[t]
	\caption{Ellipsoidal Robust OMC. Does not require simulator gradients.}\label{alg:romc_ell}
	\begin{algorithmic}[1]
		\For{$i \gets 1 \textrm{ to } n$}
		\State Obtain $\optendname$ $\optend$ and GP model distance $\gpmodel(\thetab)$ using Bayesian optimisation.
        \State Construct ellipse with volume $V_i$ using $\gpmodel(\thetab)$ as described in Subsection~\ref{subsec:prop_dist}.
		\State Define a uniform distribution $q_i(\thetab)$ over ellipse.
			\For{$j \gets 1 \textrm{ to } m$}
			\State $\thetabij \sim q_i(\thetab)$
			\State \begin{varwidth}[t]{0.9\linewidth}Accept $\thetabij$ as posterior sample with weight $w_{ij} = \indicator{\regioni}(\thetabij) * p(\thetabij) * V_i$\end{varwidth} \label{algstep:romc_ell_final_check}
			\EndFor
		\EndFor
	\end{algorithmic}
\end{algorithm}

\section{EXPERIMENTS}
\label{sec:exp}

We assess Robust OMC (ROMC) on three tasks and compare its performance to standard OMC. As reference, we use posteriors obtained by expensive Rejection ABC runs. In the appendix we further show comparison results to the exact posteriors when tractable. The accuracy is measured using the Jensen-Shannon divergence. We also contrast effective samples sizes, which are given by $\ESS=(\sum_{i=1}^n w_i)^2 / \sum_{i=1}^n w_i^2$. Note that we did not perform additional comparisons to other ABC methods because such comparisons were already performed by \citet{omc}. In their experiments, OMC showed clear improvements --- about a factor of 10 fewer calls to the simulator per accepted sample. These advantages are inherited by ROMC. %, up to additional sampling, which may incur additional cost as discussed below.

We further compared the performance of ROMC to that of two simple heuristic fixes of OMC. In the first fix, we ignored a percentage of the smallest eigenvalues of the $\jac_i\tran \jac_i$ matrices when computing the determinants, hence reducing the magnitude of the biggest weights. This is similar to using pseudo-determinants. For the second fix, we stabilised the $\jac_i\tran \jac_i$ matrices by adding a constant value to the diagonals before computing the determinant. This stabiliser value was chosen by picking a given percentile of the magnitudes of the diagonal elements of all matrices.

\subsection{Experiment 1: ROMC Resolves OMC Failure}
\label{subsec:exp_1}

We first consider a simple simulator to illustrate that ROMC resolves the identified issue of standard OMC, namely that it collapses regions of similar likelihood into a single point. The simulator is defined such that the likelihood function is flat in the area around $\xobs$ and linear otherwise: 
\begin{align}
p(x|\theta) \sim \begin{cases}
\theta^4 + u & \text{if $\theta \in [-0.5, 0.5]$} \\
\theta - c + u & \text{otherwise}
\end{cases}
\end{align}
The parameter of interest is $\theta$ and $u \sim N(0, 1)$ is a nuisance parameter and the only source of randomness. The term $c=(0.5 - 0.5^4)$ makes sure the function is continuous. Figure~\ref{fig:exp_1_example_output} in the appendix shows the simulator output for specific values of $u$. For posterior inference, we assume that the observed data is $\xobs=0$. We used Algorithm \ref{alg:romc_box} but exploited the fact that the box can be constructed analytically in this simple example.

Figure~\ref{fig:omc_fail} in the introduction shows example posteriors. Despite generating samples which span the entire range of the prior, OMC assigns much higher weights to the samples in the middle of the flat region, resulting in a posterior that is overly confident at that point. %While that point is indeed the one with the highest probability given the observed data, the points around are almost as equally likely under the true posterior, which OMC fails to capture. 
Conversely, ROMC can nearly perfectly reproduce the reference posterior. For OMC, the effective sample size divided by the number of total samples is $\ESS/n \approx 0.5$, while for ROMC, $\ESS/n \approx 0.95$. 

\begin{figure}[t]  % Experiment 1 Performances
\centering
	\centering
	\includegraphics[width=.99\linewidth]{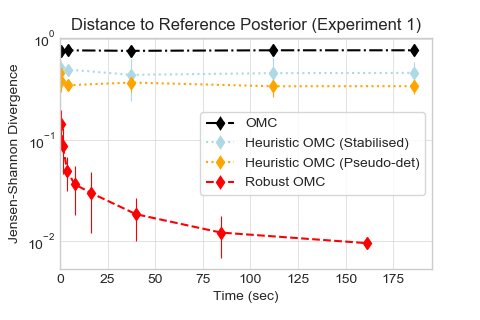}
\caption{\label{fig:exp_1_1}Experiment 1. Comparing OMC, two heuristic fixes to OMC, and Robust OMC. Smaller divergences are better.}% We compute the Jensen-Shannon divergence between the approximate posteriors and the reference Rejection ABC posterior. See Figure~\ref{fig:exp_1_true_posterior} in the appendix for a comparison against the true posterior.}
\end{figure}

Figure~\ref{fig:exp_1_1} shows how well OMC, ROMC, and the two discussed heuristic fixes of OMC can match the reference posterior at varying computational budgets. ROMC clearly outperforms OMC and both heuristics whatever the computation time. Additionally, we see that the two heuristic OMC methods perform similarly, with the pseudo-determinant version reaching a lower divergence. We will thus only consider that heuristic from now on.

\subsection{Experiment 2: ROMC as Post-processing}
\label{subsec:exp_2}

This experiment showcases that Robust OMC can be performed as post-processing to standard OMC. We assume that we can compute simulator gradients (which is required in OMC), and hence we use Algorithm~\ref{alg:romc_box}. We consider the case were the summary statistics are not completely informative about the parameters, which is a scenario that comes up often when using ABC in real-world problems \citep[e.g.][]{abc_summary_stats_aeschbacher}. As a prototypical example of this scenario, we infer the mean and variance of a normal distribution with only the sample average available as a summary statistic (see appendix for details). Since there is no direct information on the variance, the optimisation surfaces will be completely flat in one direction.

\begin{figure} % Experiment 2 Performances
	\centering
	\begin{subfigure}{.46\textwidth}
		\centering
		\includegraphics[width=.99\linewidth]{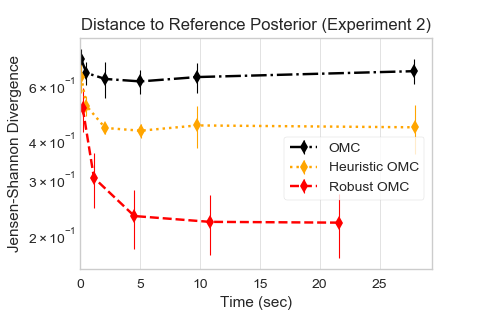}
	\end{subfigure}
	
	\caption{\label{fig:exp_2_1}Experiment 2. Comparing OMC, heuristic OMC, and Robust OMC. Smaller divergences are better.}% as a function of time taken. Like in Figure~\ref{fig:exp_1_1}, the methods are evaluated against a reference Rejection ABC posterior. See Figure~\ref{fig:exp_2_true_posterior} in the appendix for evaluation against the true posterior.}
\end{figure}

Figure~\ref{fig:exp_2_1} compares the methods for different run times against the reference Rejection ABC posterior. Heuristic OMC refers to the version based on pseudo-determinants, and we use the hyper-parameter value that produced the best result. As before, ROMC outperforms the alternatives. Figures of the posteriors themselves are shown in Figure~\ref{fig:exp_2_posteriors} in the appendix, and Heuristic OMC performance as a function of its hyper-parameter is shown in Figure~\ref{fig:exp_2_homc_hyperpar} in the appendix. 

\subsection{Experiment 3: Gradient-free ROMC}
\label{subsec:exp_3}

Here, we do not make use of the simulator gradients, and use Algorithm~\ref{alg:romc_ell}. This example is about inverse-graphics. It involves a considerably more complicated simulator that takes as input a set of 20 parameters and deterministically renders an image of a object (in this case, a teapot) on a uniform background. This is based on the generative model used by \citet{overcoming_occlusion}. We focus on the task of learning the posterior distribution of two colour parameters in a setting where there are two possible explanations for the observed image and thus the posterior is expected to be bi-modal. The remaining 18 parameters are used as nuisance parameters. They control the illumination, shape, pose and other aspects of the objects (see appendix). 

\begin{figure}
\begin{subfigure}{.25\textwidth}
		\centering
		\includegraphics[width=.6\linewidth]{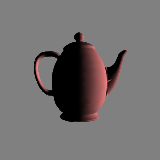}
		\caption{Gray teapot with red light.}
	\end{subfigure}%
\begin{subfigure}{.25\textwidth}
		\centering
		\includegraphics[width=.6\linewidth]{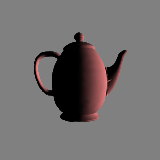}
		\caption{Red teapot with white light.}
	\end{subfigure}
	\caption{\label{fig:exp_3_teapots}Example teapots (brightened for clarity). We use \textbf{(a)} as the observed data, but \textbf{(b)} is another possible explanation.}
\end{figure}

The five illumination parameters are the most relevant ones among the nuisance variables for the task considered. The first four parameters specify the global illumination strength, the directional light strength, and the directional light angle. The fifth one allows the directional lighting to be in one of two modes: either white or red. This is what causes the bi-modality of the posterior --- if the observed image depicts a red teapot, it is both possible that it could be a grey teapot with red lighting, or a red teapot with white lighting (see Figure~\ref{fig:exp_3_teapots}). We use the former case as the observed image in our experiments.

\begin{figure*}
	\centering
	\begin{subfigure}{.333\textwidth}
		\centering
		\includegraphics[width=1.\linewidth]{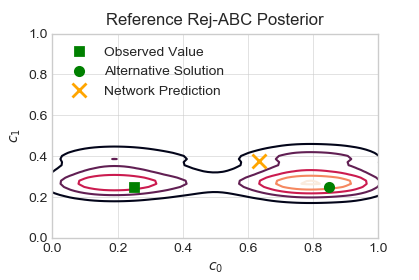}
	\end{subfigure}%
	\begin{subfigure}{.333\textwidth}
		\centering
		\includegraphics[width=1.\linewidth]{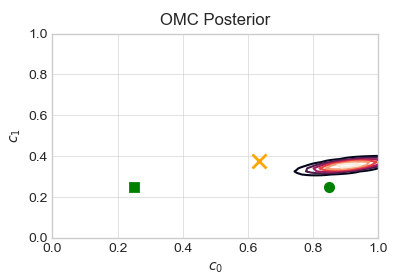}
	\end{subfigure}%
	\begin{subfigure}{.333\textwidth}
		\centering
		\includegraphics[width=1.\linewidth]{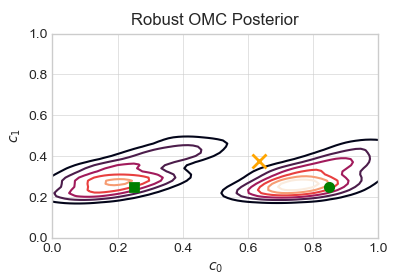}
	\end{subfigure}
	\caption{\label{fig:exp_3_post}Experiment 3 posteriors. \textbf{Left:} Reference posterior, obtained after running Rejection ABC with a high number of samples. \textbf{Middle}: Standard OMC. $\ESS/n \approx 0.005$. \textbf{Right:} ROMC with Algorithm \ref{alg:romc_ell}: $\ESS/n \approx 0.97$.}
\end{figure*}

\begin{figure}
	\centering
	\includegraphics[width=.99\linewidth]{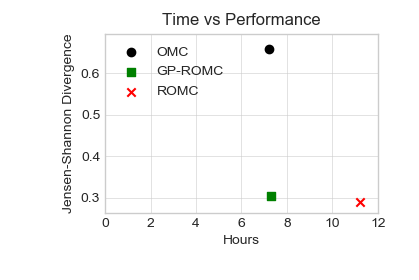}
	%\vspace{-3em}
	\caption{\label{fig:exp_3_time_vs_div}Experiment 3. Comparison between OMC and ROMC. For ROMC, we used Algorithm~\ref{alg:romc_ell} without (red cross) and with (green square) GP acceleration.}
\end{figure}

% Rotation ESS
% For OMC, $ESS/n \approx 0.009$, or $ESS = 2.2$ out of $250$ total samples.
% For ROMC, $ESS/n \approx 0.6$, or $ESS=7200$ out of $12000$ accepted samples.
% Colour ESS
% For OMC, $ESS/n \approx 0.005$, or $ESS = 1.2$ out of $250$ total samples.
% For ROMC, $ESS/n \approx 0.97$, or $ESS=24300$ out if $25000$ total samples.

The generative model was implemented using Open Differential Renderer \citep[OpenDR, ][]{opendr}. While this renderer does allow us to compute the simulator gradients, we did not use them for any of our presented results. To compute the distance $d(\Xsim, \Xobs)$ between a simulated image $\Xsim = \gen(\thetab, \nuisance_i)$ and the observed image $\Xobs$, we use a recognition model that predicts the target parameters $\thetab$ for an input image $\X$ and then compute the Euclidean distance between the two images' predicted parameters, so that
\begin{align}
d(\Xsim, \Xobs) = \sqrt{\sum_{n=1}^{|\thetab|} (r_n(\Xsim) - r_n(\Xobs)})^2,
\end{align}
where $r_n(\cdot)$ is the prediction of the $n$-th parameter. The parameter estimates can thus be viewed as summary statistics. We implemented the recognition model as a neural network that was pre-trained on  data generated from the simulator using a broad prior on the nuisance variables under daylight (white illumination). For details about the neural network's architecture, training procedure, and performance, as well as for examples of the training data, see the appendix.

Obtaining a single $\optendname$ $\optend$, i.e.\ minimising the distance $d(\Xsim, \Xobs)$ with respect to the colour parameters, took approximately 2 minutes. We ran all simulations on a single computer only, and did not exploit the possibility to parallelise the inference. We based our posterior approximation on 250 $\optendname$s. For ROMC, we generated 100 new samples per original $\optendname$. %yielding a total of 25000 posterior samples.

The OMC and ROMC posteriors are shown in Figure~\ref{fig:exp_3_post}, along with a Rejection ABC reference posterior and the predictions by the recognition network (see the appendix for results with the pseudo-determinant heuristic). First of all, we see that the recognition network prediction is off even though the network was well trained (see Figure \ref{fig:exp_3_nns} in the appendix). This is because red lightning conditions were not part of the training data and the recognition network does not well generalise towards this condition. Indeed, while still not accurate, the recognition network favours the solution in Figure \ref{fig:exp_3_teapots}(b), which is closer to images typically seen during training. Among the Bayesian methods, OMC is overly confident at a single location, with an effective sample size of $1.2$ (out of 250 total samples). What is more, its posterior is not centred on a viable solution. On the other hand, ROMC produces two distinct posterior modes that contain the two possible solutions as we would want in this scenario, and it generally matches the reference posterior. Remarkably, these results were obtained despite using a biased recognition network, which points to a general ability of ABC in dealing with systematic biases in recognition networks.

Figure~\ref{fig:exp_3_time_vs_div} compares the trade-off between accuracy and compute time for OMC (black) and ROMC (red and green). We see that ROMC as implemented in Algorithm~\ref{alg:romc_ell} (red) is much more accurate than OMC but that it incurs an extra cost. This extra cost is due to the additional runs of the simulator that are needed for the acceptance check $\indicator{\regioni}(\thetabij)$ in step~\ref{algstep:romc_ell_final_check} of the algorithm. This cost could be reduced by parallelising the runs (which we did not do). Alternatively, we can use the GP model of the distance rather than the true distance in the acceptance check, which does not require additional runs of the simulator. We call this approach GP-ROMC. The figure shows that GP-ROMC (green) has almost the same accuracy as ROMC, but that it is much faster and only incurs a tiny overhead compared to OMC. The posterior for GP-ROMC is shown in Figure~\ref{fig:exp_3_post_gpromc} in the appendix. In line with the numerical result, the posterior is very close to one obtained with exact acceptance checks.

%%%%%%%%%%%%%%%%%%%%%%%%%%%%%%%%%%%%%%%%%%%%%%%%%%%%%%%%%%%%
%%%%%%%%%%%%%%%%%%%%%%% Conclusion %%%%%%%%%%%%%%%%%%%%%%%
%%%%%%%%%%%%%%%%%%%%%%%%%%%%%%%%%%%%%%%%%%%%%%%%%%%%%%%%%%%%
\section{CONCLUSIONS}
\label{sec:conc}
\vspace{-0.3em}
This paper dealt with the task of performing Bayesian inference for parametric models when the likelihood is intractable but sampling from the model is possible. We considered Optimisation Monte Carlo (OMC) which has been shown to be a promising tool to efficiently sample from an approximate posterior. We showed that OMC, while efficient, has the shortcoming that it collapses regions of similar or near-constant likelihood into a single point. This matters because OMC samples might thus severely under-represent the uncertainty in the posterior and hence produce overly confident predictions. 

We addressed this issue by introducing the more general framework of Robust Optimisation Monte Carlo (ROMC) and two concrete algorithms implementing it.
The ROMC framework can be considered to be a form of ABC where we use optimisation to automatically construct suitable and localised proposal distributions. The first algorithm can be run as a form of post-processing after standard OMC to correct for the identified pathology. The second algorithm, unlike OMC, can be used when (approximate) gradients are not available. It uses a surrogate model of the distance and we have seen that this approach can be used to almost entirely eliminate the extra cost of ROMC compared to OMC. It is hence reasonable to also use a surrogate model in the first algorithm if reducing compute cost is necessary. 

We tested the proposed framework and algorithms on both prototypical toy examples and complex inference tasks from inverse-graphics, and found that the proposed ROMC approach did accurately estimate the posteriors while OMC did not. 

\vspace{-0.25em}
\section*{Acknowledgements}
\vspace{-0.5em}

This work was supported in part by the EPSRC Centre for Doctoral Training in Data Science, funded by the UK Engineering and Physical Sciences Research Council (grant EP/L016427/1) and the University of Edinburgh.
% \newpage

%\clearpage
\bibliographystyle{plainnat}
\bibliography{ms}

%%%%%%%%%%%%%%%%%%%%%%%%%%%%%%%%%%%%%%%%%%%%%%%%%%%%%%%%%%%%
%%%%%%%%%%%%%%%%%%%%%%% Appendix %%%%%%%%%%%%%%%%%%%%%%%
%%%%%%%%%%%%%%%%%%%%%%%%%%%%%%%%%%%%%%%%%%%%%%%%%%%%%%%%%%%%
\vspace{3em}
\appendix
{\Large{\textbf{Appendix}}}

%------------------------------
\section{Rejection ABC Algorithm}
\begin{algorithm}
	\caption{Rejection ABC. Generates $n$ independent samples $\thetab_i$ of the approximate posterior. Needs black-box simulator $\gen(\thetab, \nuisance)$, observed data $\Xobs$, computational budget $N$, and number of accepted samples $n$.}\label{alg:rej_abc}
	\begin{algorithmic}[1]
		\For{$i \gets 1 \textrm{ to } N$}
		\State $\thetab_i \sim p(\mathbf{\thetab})$ \Comment{Draw parameters $\thetab_i$ from prior.}
		\State $\Xsim \sim \gen(\thetab_i,\cdot)$ \Comment{Simulate synthetic data using $\thetab_i$.}
		\State $d_i = d(\Xsim, \Xobs)$ \Comment{Compute distance to $\Xobs$.}
		\EndFor
		\State Accept the $n$ samples $\thetab_i$ with the lowest distance $d_i$ as samples from the posterior.
	\end{algorithmic}
\end{algorithm}

A standard variant of Rejection ABC accepts only those samples whose distance falls under a threshold $\epsilon$ specified by the user.

%------------------------------
\section{Proof of Theorem~\ref{theo:omc}}

As a reminder, the assumptions under which ROMC becomes standard OMC as $\epsilon \to 0$ are:

\textbf{Assumption 1.} The distance $d(\gen(\thetab, \nuisance), \Xobs)$ is given by the Euclidean distance between summary statistics $||\gensum(\thetab, \nuisance) - \sumstatObs||$.

\textbf{Assumption 2.} The proposal distribution $q_i(\thetab)$ is the uniform distribution on $\regioni$.

\textbf{Assumption 3.} The acceptance regions $\regioni$ are approximated by the ellipsoid $\regioni = \{\thetab : (\thetab - \optend)\tran \jac_i\tran\jac_i (\thetab - \optend) \leq \epsilon\}$ where $\jac_i$ is the Jacobian matrix with columns $\partial \gensum(\optend, \nuisance_i) / \partial \theta_k$.

\textbf{Assumption 4.} The matrix square root $\A_i$ of $\jac_i\tran\jac_i$ is full rank, i.e.\ $\text{rank}(\A_i)=\text{dim}(\thetab)$. 

\textbf{Assumption 5.} The prior is constant on the acceptance regions $\regioni$, i.e.\ $p(\thetab) = p(\optend)$ for $\thetab \in \regioni$. 

Since settings of $\nuisance_i$ that result in empty sets $\regioni$ (i.e.\ no $\optend$ exists such that the resulting distance is below $\epsilon$) are excluded from affecting the approximate posterior (both in ROMC and in standard OMC), we here consider only the case of non-empty sets $\regioni$. 

We start from Equation \eqref{eq:romc_eff} using for $q_i(\thetab)$ --- as per Assumption 2 --- the uniform distribution on $\regioni$ with density $\regionuni$,
\begin{equation}
    \regionuni = \frac{1}{\voli} \indicator{\regioni}(\thetab).
\end{equation}
Since the proposal distribution $q_i(\thetab)$ is zero outside $\regioni$, we have $\indicator{\regioni}(\thetabij)=1$ for all $\thetabij$ and hence
\begin{align}
    \E[h(\thetab) | \Xobs]  \approx&  \frac{ \sum_{i=1}^{n} \sum_{j=1}^{m} h(\thetabij)\voli p(\thetabij)}{\sum_{i=1}^{n} \sum_{j=1}^{m} \voli p(\thetabij)}.
\end{align}
By Assumption 3, $\regioni$ is an ellipsoid with volume determined by the matrix square root $\A$ of $\jac_i\tran\jac_i$, as well as the value of $\epsilon$. It is possible to split the volume into a term determined by the shape of the ellipsoid and a term determined by $\epsilon$. With the change of variables $\w = \A_i(\thetab-\optend)$, we have:
\begin{align}
\voli &= \int_{\regioni}\ d\thetab = \int_{\mathbf{w}:||\mathbf{w}||^2\le\epsilon} \determ \ d \w \\
&= \determ \vole,
\end{align}
where $B_{\epsilon}$ denotes an $\epsilon$-ball in Euclidean space. By Assumption 4, $\determ$ is finite. Low-rank matrices $\A_i$ would correspond to ellipsoids that extend without bound into one (or more) directions. We thus obtain
\begin{align}
    \hspace{-5ex}\E[h(\thetab) | \Xobs]  \approx&  \frac{ \sum_{i=1}^{n} \sum_{j=1}^{m} h(\thetabij) \frac{\vole}{\dete} p(\thetabij)}{\sum_{i=1}^{n} \sum_{j=1}^{m} \frac{\vole}{\dete} p(\thetabij)}\\
    \approx&  \frac{ \sum_{i=1}^{n}  \determ \sum_{j=1}^{m} h(\thetabij)p(\thetabij)}{\sum_{i=1}^{n}  \determ \sum_{j=1}^{m} p(\thetabij)}
\end{align}
where we cancelled $\vole$ so that only the term $\determ$ reflecting the geometry of the ellipsoid remains. Note that $\determ$ can also be written as $\determ = (\det \jac_i\tran\jac_i)^{-1/2}$. By Assumption 5, $p(\thetabij)=p(\optend)$, so that we have
\begin{align}
    \E[h(\thetab) | \Xobs]  \approx&  \frac{ \sum_{i=1}^{n} (\det \jac_i\tran\jac_i)^{-\frac{1}{2}} p(\optend) \sum_{j=1}^{m} h(\thetabij)}{\sum_{i=1}^{n} (\det \jac_i\tran\jac_i)^{-\frac{1}{2}} p(\optend)}
\end{align} 
In this expression, the only dependency on $\epsilon$ remains in the samples $\thetabij \sim \regionuni$. In the limit of infinitely small $\epsilon$, $\regionuni$ becomes a Dirac delta distribution $\delta(\thetab-\optend)$ centred at $\optend$. This means that the only possible sample from that distribution is $\optend$ and hence that $h(\thetabij) = h(\optend)$ for all $j$. In the limit of $\epsilon \to 0$, we thus obtain
\begin{align}
    \E[h(\thetab) | \Xobs]  \approx&  \frac{ \sum_{i=1}^{n} (\det \jac_i\tran\jac_i)^{-\frac{1}{2}} p(\optend) h(\optend)}{\sum_{i=1}^{n} (\det \jac_i\tran\jac_i)^{-\frac{1}{2}} p(\optend)}.
\end{align} 
This expression is a weighted average using samples $\optend$ and weights $w_i$ as defined in Algorithm \ref{alg:omc}. This means that the stated assumptions yield the weighted posterior samples of OMC and concludes the proof.

%------------------------------
\section{Constructing the Proposal Region for ROMC}

Here we give more details on how we obtain the proposal distribution $q_i(\thetab)$, using Figure~\ref{fig:implementation_example} as a visual aid. We start with the optimisation end point $\optend$ (blue cross in Figure~\ref{fig:implementation_example}). We also have the curvature matrix, which is $\jac_i\tran \jac_i$ in Algorithm~\ref{alg:romc_box} or the Hessian of the GP model at $\optend$ in Algorithm~\ref{alg:romc_ell}. Its eigenvectors are used to determine what we call scan directions --- one per dimension, along with its opposite (orange lines). We move along these directions until the resulting distance to $\Xobs$ is no longer under a certain threshold. The distance is calculated with the GP model if it is available in order to speed this process up. These end points along the scan directions are then used to create a \textit{loose} rectangular box. This procedure is summarised in Algorithm~\ref{alg:region_construction}. The loose box is then either the final proposal region (Algorithm~\ref{alg:romc_box}), or is used to draw training data for the regression model to create an ellipse which is the final proposal region (Algorithm~\ref{alg:romc_ell}). In both cases, placing a uniform distribution on this region gives the final proposal distribution $q_i(\thetab)$.

While we do use $\epsilon$ for the final $\indicator{\regioni}(\thetabij)$ check, it would be reasonable to make the proposal region slightly bigger in order to ensure we capture as much of the actual acceptance region as possible at the cost of rejecting a few more samples. We achieve this by specifying the threshold for finding $q_i(\thetab)$ to be bigger than the threshold used for $\indicator{\regioni}(\thetabij)$. In Algorithm~\ref{alg:romc_box}, we use $\epsilon_{prop}$ as defined by the 95\% quantile of the $\optendname$ distances to define the proposal region. In Algorithm~\ref{alg:romc_ell}, we use $\epsilon_{prop}$ based on the same 95\% quantile on the ellipse, and a threshold based on a bigger 97.5\% quantile to define the loose box from which the training data for the regression model is drawn. It is important to note that just using the single 90\% quantile for all thresholds still produces a good final posterior, so the method is to some extent robust to that choice. The bigger thresholds we propose above produce a very slight performance improvement in practice (the divergence to the reference posterior with the bigger thresholds is about 1\% smaller,\footnote{The exact numbers for the Jensen-Shannon divergences are 0.290 and 0.293.} which indicates a better performance) and also make intuitive sense, which is why we use them in our final implementation. 

For Algorithm~\ref{alg:romc_ell}, the regression model is quadratic and is trained via least squares. As it is quadratic, its contours are ellipsoidal. Thus, by finding the contour equal to $\epsilon_{prop}$, we obtain an ellipse that is suitable for being the proposal region.

\begin{algorithm}[t]
	\caption{Box construction for one iteration $i$ of Algorithms~\ref{alg:romc_box} and~\ref{alg:romc_ell}. Needs $\optend$, $\nuisance_i$, step size $\eta$, number of refinements $K$, and curvature matrix $\hessian_i$ ($\jac_i\tran\jac_i$ if Algorithm~\ref{alg:romc_box} or GP Hessian if Algorithm~\ref{alg:romc_ell}).}\label{alg:region_construction}
	\begin{algorithmic}[1]
	\State Compute eigenvectors $\mathbf{v}_{dim}$ of $\hessian_i$ {\scriptsize ($dim = 1,\ldots,||\thetab||)$}
	\For{$dim \gets 1 \textrm{ to } ||\thetab||$}
		\State $\Tilde{\thetab} \gets \optend$ \label{algstep:box_constr_start}
		\State $k \gets 0$
		\Repeat
        	\Repeat
                \State $\Tilde{\thetab} \gets \Tilde{\thetab} + \eta \ \mathbf{v}_{dim}$ \Comment{Large step size $\eta$.}
        	\Until{$d(\gen(\Tilde{\theta}, \nuisance_i), \Xobs) \ge \epsilon$}
        	\State $\Tilde{\thetab} \gets \Tilde{\thetab} - \eta \ \mathbf{v}_{dim}$
        	\State $\eta \gets \eta/2$ \Comment{More accurate region boundary\footnotemark.}
        	\State $k \gets k + 1$
    	\Until $k = K$
    	\State Set final $\Tilde{\thetab}$ as region end point. \label{algstep:box_constr_end}
    	\State Repeat steps \ref{algstep:box_constr_start}~-~\ref{algstep:box_constr_end} for $\mathbf{v}_{dim} = - \mathbf{v}_{dim}$
	\EndFor
	\State Fit a rectangular box around the region end points.
	\end{algorithmic}
\end{algorithm}

Additionally, to show that Algorithm~\ref{alg:romc_ell} is reasonably robust to the construction of the loose box, we compare the above proposal construction method with an alternative one. This additional method works as follows: begin with a tiny box centred at the optimisation end point, aligned with the scan directions obtained from the GP model's Hessian. Sample $\thetab$ values uniformly from the box, compute their distance using the GP model, and check how many are under the acceptance threshold. Gradually expand this box until 50\% of the sampled points are no longer within the threshold. That is the final loose region, which is then used to train the regression model and thus to produce an ellipse as before. \footnotetext{{This provides a minor benefit and is not necessary.}}The divergence between the resulting final posterior and the reference posterior is less than 3\% bigger than when we use Algorithm~\ref{alg:romc_ell} as described previously.\footnotemark This difference is quite small, implying some robustness to the minute details of the construction of the proposal region.

\footnotetext{The exact numbers for the Jensen-Shannon divergences are 0.290 and 0.298.}

\begin{figure*}[ht] %%%%% Exp 1: Underlying Functions %%%%%
	\begin{subfigure}{0.33\textwidth}
		\centering
		\includegraphics[width=1.\linewidth]{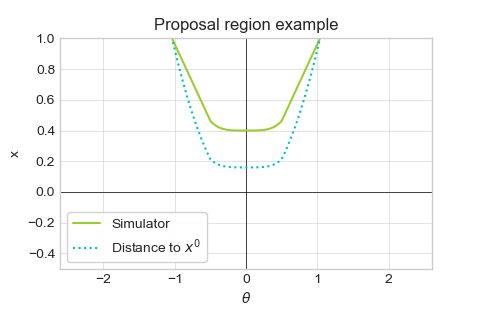}
		\caption{$u=0.4$}
	\end{subfigure}
	\begin{subfigure}{0.33\textwidth}
		\centering
		\includegraphics[width=1.\linewidth]{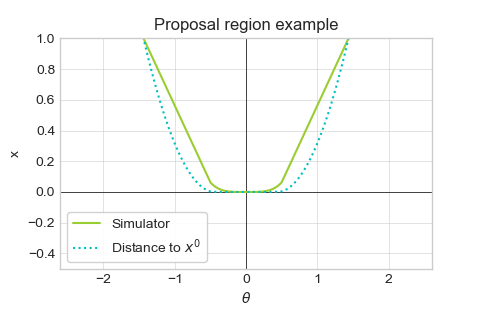}
		\caption{$u=0$}
	\end{subfigure}
	\begin{subfigure}{0.33\textwidth}
		\centering
		\includegraphics[width=1.\linewidth]{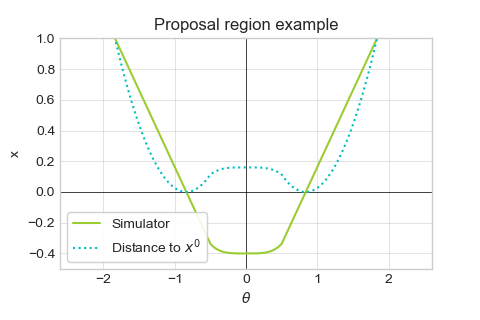}
		\caption{$u=-0.4$}
	\end{subfigure}
	\caption{\label{fig:exp_1_example_output}Experiment 1. Examples of the simulator's output and its distance to the observed data $\xobs=0$ for three specific values of the nuisance parameter $u$.}
\end{figure*}

\begin{figure*}[ht]
	\begin{minipage}[t]{0.45\textwidth} %%%%% Exp 1: JS-vs-Time %%%%%
		\centering
		\includegraphics[width=1.\linewidth]{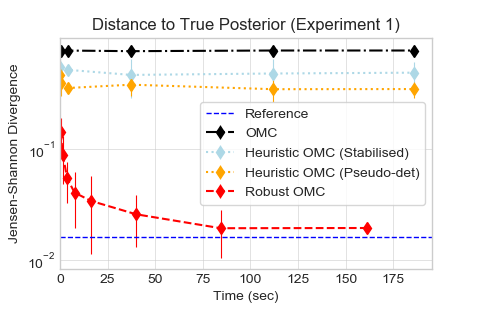}
		\caption{\label{fig:exp_1_true_posterior}Experiment 1. Comparison of OMC, Robust OMC, and the two heuristic OMC methods against the true posterior. The reference Rejection ABC posterior's divergence is also shown.}
	\end{minipage}\hfill
	\begin{minipage}[t]{0.45\textwidth} %%%%% Exp 2: JS-vs-Time %%%%%
		\centering
		\includegraphics[width=1.\linewidth]{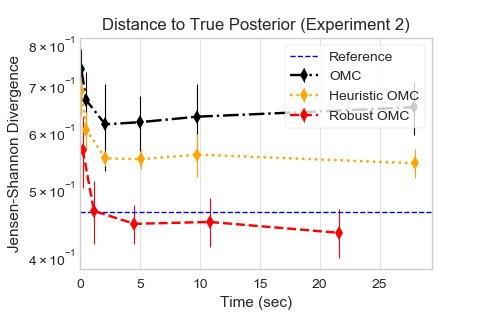}
		\caption{\label{fig:exp_2_true_posterior}Experiment 2. Comparison of OMC and Robust OMC against the true posterior. The reference posterior's divergence is also shown.}
	\end{minipage}
\end{figure*}

%------------------------------
\section{Additional Information for Exp.\ 1}
\vspace{-0.4em}

In this section, we further discuss Experiment 1 from Subsection~\ref{subsec:exp_1}. As a reminder, the likelihood function we use is defined by: 
\begin{align}
p(x|\theta) \sim \begin{cases}
\theta^4 + u & \text{if $\theta \in [-0.5, 0.5]$} \\
\theta - c + u & \text{otherwise}
\end{cases}
\end{align}
where $u \sim N(0, 1)$ is a nuisance parameter and the only source of randomness, and the term $c=(0.5 - 0.5^4)$ ensures the function is continuous. We assume that the observed data is $\xobs=0$ and that the distance is Euclidean. 

Figure~\ref{fig:exp_1_example_output} shows the simulator output for specific values of $u$. In the case of $u>0$, the simulator can never generate a data point that matches $\xobs=0$ for any $\theta$, although for $u$ sufficiently close to $0$, some $\theta$ may result in a data point within the distance threshold.

For $u<0$, we enter the interesting situation where there are two values for which the distance is $0$, and the distance is non-zero between them. In other words, there are two possible zero-solutions to the OMC optimisation objective $d(\gen(\theta, u_i), \xobs)$. This would imply that there can be two disjointed acceptance regions $\regioni$ for a single $\theta^*_i$ if the threshold $\epsilon$ is small enough. Currently, both OMC and our two Robust OMC implementations would not be able to capture the full disjointed region $\regioni$ in such a scenario --- OMC would at best approximate only the size of the region around $\theta^*_i$ with a weight, and Robust OMC would construct the box / ellipse and hence the proposal distribution only around $\theta^*_i$ as well. This problem does not manifest in the results we have presented as we computed the correct acceptance region analytically. In general, this is a difficult issue to solve, although a simple fix (i.e.\ a potential improvement on Robust OMC) would be to restart the optimiser at different initial values in order to find all possible solutions to the optimisation objective. It is also possible that, with enough samples, errors from the disjointed regions would average out and thus the final posterior would still be correct.

In the main text, Figure~\ref{fig:exp_1_1} compared OMC, Robust OMC, and two heuristic OMC methods against a reference posterior obtained from an expensive Rejection ABC run. Here we show the comparison made against the true posterior (which can be computed analytically) in Figure~\ref{fig:exp_1_true_posterior}. As before, Robust OMC outperforms the other methods. It does not quite reach the level of the reference Rejection ABC method (blue dashed line) but this is to be expected as the reference was ran for a much longer time than the other methods.

%------------------------------
\vspace{-0.4em}
\section{Additional Information for Exp. 2}
\vspace{-0.4em}
Here we present further details and results for Experiment 2 discussed in Subsection~\ref{subsec:exp_2}. Recall that the task was about inferring the parameters of a 2D Gaussian with a non-informative summary statistic, namely just the mean of a sample from the Gaussian, while also having access to the simulator gradients.

We assume that the sample size is $M=25$, and that the observed sample average is $\mu^0=1$. We chose a Gaussian prior for the mean and Inverse-Gamma prior for the standard deviation: $p(\mu) = \mathcal{N}(0, 5)$, $p(\sigma) = \text{Inv-Gamma}(0.2, 1)$. To compute the exact posterior, we used the fact that for a sample of size $M$ from a normal $\mathcal{N}(\mu, \sigma^2)$ the sample mean is distributed according to $\mathcal{N}(\mu, \sigma^2/M)$.

\begin{figure*}[ht] %%%%% Exp 2: Posteriors %%%%%
	\centering
	\makebox[\textwidth][c]{
		\begin{subfigure}{0.8\textwidth}
			\centering
			\includegraphics[width=1.\linewidth, trim={9.7cm 0 2.6cm 0},clip]{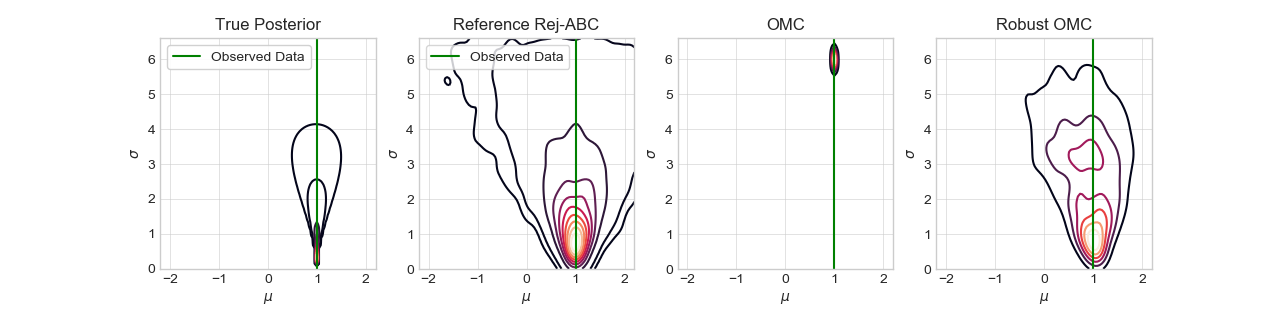}
		\end{subfigure}
	}
	\caption{\label{fig:exp_2_posteriors}Experiment 2 posteriors. For OMC, $\ESS/n \approx 0.02$, implying that the vast majority of the samples are ignored. For Robust OMC, $\ESS/n \approx 0.55$, which is a significant improvement.}
\end{figure*}

In Figure~\ref{fig:exp_2_true_posterior}, we show the performance of the methods compared against the true posterior rather than the reference ABC one used in the main text. The Jensen-Shannon divergence between Robust OMC's posterior and the true posterior is much lower than for the other methods. The actual posteriors themselves are shown in Figure~\ref{fig:exp_2_posteriors}. OMC correctly identifies the marginal over $\mu$ but fails to do so for $\sigma$ and does not match the reference posterior. On the other hand, Robust OMC reasonably matches the reference. Also note that OMC's effective sample size divided by the number of total samples is $\ESS/n \approx 0.01$, implying that the vast majority of the samples are ignored. Conversely, the corresponding value for Robust OMC is $\ESS/n \approx 0.55$, which is a significant improvement.

Figure~\ref{fig:exp_2_homc_hyperpar} additionally compares the performance of our pseudo-determinant heuristic fix to OMC for different values of the hyper-parameter. This hyper-parameter represents the number of eigenvalues ignored when computing the determinants of the $\jac_i\tran \jac_i$ matrices and hence the weights. While the divergence to the reference posterior does change, there is still a large gap between the best heuristic OMC and the robust OMC result.

%------------------------------
\vspace{-0.4em}
\section{Additional Information for Exp. 3}
\vspace{-0.4em}

\begin{figure*} %%%%% GP-ROMC posterior
	\centering
	\begin{subfigure}{.333\textwidth}
		\centering
		\includegraphics[width=1.\linewidth]{Figures/Task_3/Colour/rej_abc_bivariate_c0_c1.png}
	\end{subfigure}%
	\begin{subfigure}{.333\textwidth}
		\centering
		\includegraphics[width=1.\linewidth]{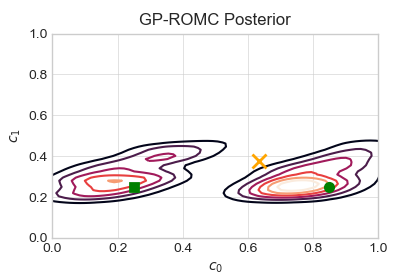}
	\end{subfigure}%
	\begin{subfigure}{.333\textwidth}
		\centering
		\includegraphics[width=1.\linewidth]{Figures/Task_3/Colour/posterior_bivariate_romc_c0_c1.png}
	\end{subfigure}
	\caption{\label{fig:exp_3_post_gpromc}Experiment 3 posteriors. Similar to Figure~\ref{fig:exp_3_post}, but with the GP-ROMC approach (middle) shown for comparison as well. }
\end{figure*}

\begin{figure*} %%%%% Exp 3: Heuristic OMC %%%%%
	\centering
	\begin{minipage}[t]{0.45\textwidth} 
		\centering
		\includegraphics[width=1.\linewidth]{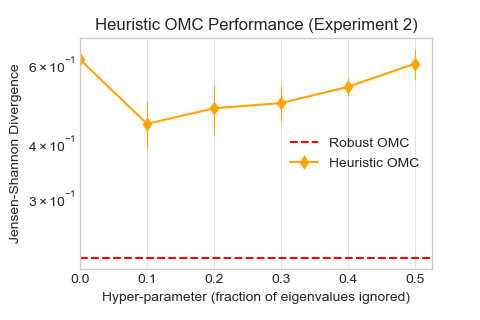}
		\caption{\label{fig:exp_2_homc_hyperpar}Performance of the heuristic pseudo-determinant fix to OMC as a function of its hyper-parameter in Experiment 2, compared against Robust OMC ran with roughly the same computational budget.}
	\end{minipage}\hfill
	\begin{minipage}[t]{0.45\textwidth} %%%%% Exp 2: JS-vs-Time %%%%%
		\centering
		\includegraphics[width=1.\linewidth]{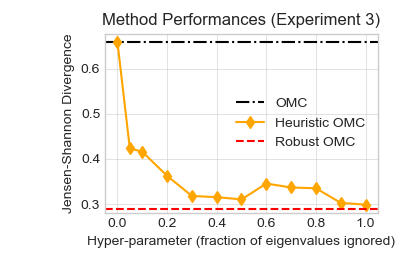}
		\caption{\label{fig:exp_3_performances}Performance of the heuristic pseudo-determinant fix to OMC in Experiment 3, compared against OMC and Robust OMC.}
	\end{minipage}
\end{figure*}

\textbf{Additional results.} Figure~\ref{fig:exp_3_post_gpromc} qualitatively compares GP-ROMC --- the approach where we speed up the final distance check in Algorithm~\ref{alg:romc_ell} by using the GP model instead of the simulator --- to the standard Robust OMC approach and the reference posterior. Additionally, Figure~\ref{fig:exp_3_performances} compares overall Robust OMC performance to that of OMC and Heuristic OMC. Comparing Robust OMC to Heuristic OMC, we noticed that as more eigenvalues are ignored for Heuristic OMC, the weights become more similar and the accuracy of the posterior improves. This is because, in this particular example and unlike before, the unweighted samples $\optend$ do reasonably represent the posterior so that setting all weights to a constant provides a reasonable solution. However, such tuning is not possible in practice where a reference posterior is not available.

\textbf{All renderer parameters.} The full list of parameters we use for the renderer in Experiment 3 in Subsection~\ref{subsec:exp_3} is as follows:

\begin{itemize}
    \item Ten shape parameters. The object's exact shape is based on a morphable mesh specified by Principal Component Analysis. The 10 dimensions used are the 10 highest principal components.
    \item Two rotation parameters, specifically the azimuth and elevation. The camera is always centred at the midpoint of the object.
    \item Three colour parameters --- an RGB array which globally identifies the colour of the object. The first two (the red and green channels) are the target parameters $\thetab$ over which we perform inference in Experiment 3.
    \item Five illumination parameters that characterise the lighting on the object. Unlike \citet{overcoming_occlusion} who use spherical harmonics to model illumination, we use single-source directional lighting as it is more intuitive and natural.
\end{itemize}

\begin{figure*}[ht] %%%%% Exp 3: Dataset samples %%%%%
	\centering
	\begin{subfigure}{0.16\textwidth}
		\includegraphics[width=1.\linewidth]{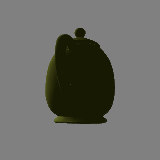}
	\end{subfigure}
	\begin{subfigure}{0.16\textwidth}
		\includegraphics[width=1.\linewidth]{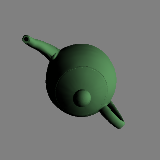}
	\end{subfigure}
	\begin{subfigure}{0.16\textwidth}
		\includegraphics[width=1.\linewidth]{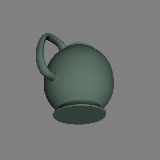}
	\end{subfigure}
	\begin{subfigure}{0.16\textwidth}
		\includegraphics[width=1.\linewidth]{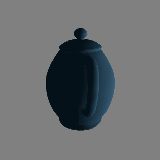}
	\end{subfigure}
	\begin{subfigure}{0.16\textwidth}
		\includegraphics[width=1.\linewidth]{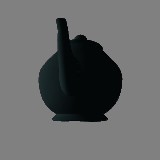}
	\end{subfigure}
	\begin{subfigure}{0.16\textwidth}
		\includegraphics[width=1.\linewidth]{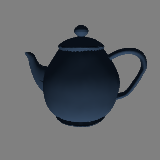}
	\end{subfigure}
		
	\begin{subfigure}{0.16\textwidth}
		\includegraphics[width=1.\linewidth]{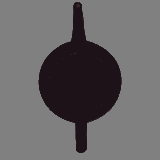}
	\end{subfigure}
	\begin{subfigure}{0.16\textwidth}
		\includegraphics[width=1.\linewidth]{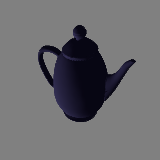}
	\end{subfigure}
	\begin{subfigure}{0.16\textwidth}
		\includegraphics[width=1.\linewidth]{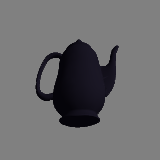}
	\end{subfigure}
	\begin{subfigure}{0.16\textwidth}
		\includegraphics[width=1.\linewidth]{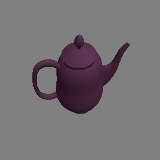}
	\end{subfigure}
	\begin{subfigure}{0.16\textwidth}
		\includegraphics[width=1.\linewidth]{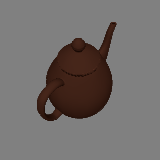}
	\end{subfigure}
	\begin{subfigure}{0.16\textwidth}
		\includegraphics[width=1.\linewidth]{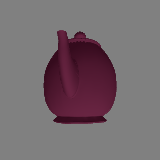}
	\end{subfigure}
	\caption{\label{fig:exp_3_data_samples}Experiment 3. Examples from the training set used to train the recognition network.}
\end{figure*}

\begin{figure*} %%%%% Exp 3: Network performance %%%%%
	\centering
	\begin{subfigure}{.4\textwidth}
		\centering
		\includegraphics[width=.99\linewidth, trim={0 0 0 0},clip]{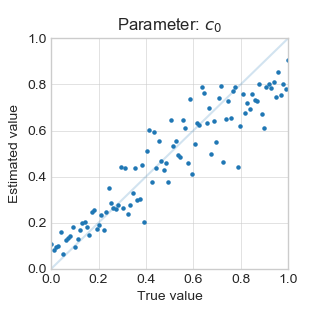}
	\end{subfigure}%
    \begin{subfigure}{.4\textwidth}
		\centering
		\includegraphics[width=.99\linewidth, trim={0 0 0 0},clip]{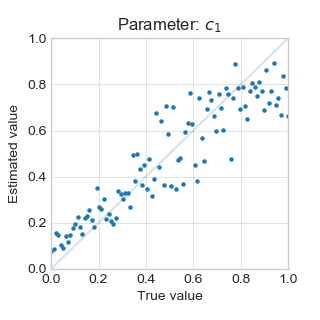}
	\end{subfigure}
	\caption{\label{fig:exp_3_nns}Experiment 3. Neural network predictions for the first two colour parameters $c_0$, and $c_1$. Line where true value is exactly equal to the predicted value is given for reference.}
\end{figure*}

\textbf{Recognition network details.} The network we used has 3 convolutional layers, each with 64 5x5 filters and 2x2 max pooling, followed by 2 linear layers with 256 and 64 hidden units respectively. Each layer uses ReLU activation functions except the final layer which uses an identity activation. The network parameters were learned with Adagrad \citep{adagrad} as it showed the most robustness to the values of the hyper-parameters of the neural network training procedure (batch size, learning rate, and dropout probability), which in turn were chosen via hyper-parameter optimisation. Additionally, Figure~\ref{fig:exp_3_data_samples} shows samples from the training set used to train the recognition network. Note that there is a reasonable amount of variability in shape, pose, illumination, and colour. Figure~\ref{fig:exp_3_nns} shows that the learned recognition network is reasonably good at reconstructing the parameters for test images from the training data.

\textbf{Gaussian process model and Bayesian optimisation details.} For simplicity, we used the default options in GPyOpt when running our experiments --- namely the Expected Improvement acquisition function, the Matern 5/2 kernel, 50 optimisation iterations, and 5 optimisation restarts. Importantly, experimentation with different settings did not lead to noticeable differences. We should also note that while using the GP model instead of the simulator in Experiment 3 incurred a significant speed up, the benefits of the GP model largely depend on the simulator cost. However, in the second experiment (where the simulator is very fast) using the GP model is still about as fast as using the simulator itself, and both approaches produce good results.

% ------------------- Figures ---------------------------

\end{document}